\documentclass[lettersize,journal]{IEEEtran}
\usepackage{amsmath,amsfonts}
\usepackage{algorithmic}
\usepackage{array}
\usepackage[caption=false,font=normalsize,labelfont=sf,textfont=sf]{subfig}
\usepackage{textcomp}
\usepackage{stfloats}
\usepackage{url}
\usepackage{verbatim}
\usepackage{graphicx}
\def\BibTeX{{\rm B\kern-.05em{\sc i\kern-.025em b}\kern-.08em
    T\kern-.1667em\lower.7ex\hbox{E}\kern-.125emX}}
\usepackage{balance}

\usepackage{booktabs}
\usepackage{multirow}
\usepackage[table]{xcolor}
\usepackage{pifont}
\usepackage{subcaption}
\usepackage{makecell}

\begin{document}

\title{Stabilizing the Modality Gap: Null-Space Projected Continual Learning for Vision–Language Models}

\author{
Tiantian Peng,
Yuyang Liu*,
Shuo Yang,
Qiuhe Hong,
Yonghong Tian*%
\thanks{This work has been submitted to the IEEE for possible publication. Copyright may be transferred without notice, after which this version may no longer be accessible.}%
\thanks{Tiantian Peng, Shuo Yang, and Qiuhe Hong are with the School of Electronic and Computer Engineering, Peking University Shenzhen Graduate School, Shenzhen 518055, China (e-mail: meetforever@stu.pku.edu.cn; shuo\_yang@stu.pku.edu.cn; lluosi@stu.pku.edu.cn).}%
\thanks{Yuyang Liu and Yonghong Tian are with the Guangdong Provincial Key Laboratory of AI for Science and Computing (AI4SC), School of Artificial Intelligence for Science, Peking University Shenzhen Graduate School, Shenzhen 518055, China (e-mail: liuyuyang13@pku.edu.cn; yhtian@pku.edu.cn).}%
\thanks{Corresponding authors: Yuyang Liu and Yonghong Tian.}%
}

\maketitle

\begin{abstract}
The modality gap is intrinsic to multimodal alignment, where embeddings from different modalities remain partially separated in a shared representation space. In continual learning, CLIP-based vision–language models (VLMs) not only suffer catastrophic forgetting but also experience modality gap drift that disrupts the shared embedding geometry. Such drift degrades cross-modal alignment and generalization.
We propose FPCD, an efficient framework combining Fast Null-space Projection (FNP) and Contrastive Distillation (CD). FNP accumulates compact Gram matrices of previous tasks activations and projects gradients onto layer-adaptive approximate null spaces to reduce interference with learned representations. By replacing singular value decomposition with symmetric eigendecomposition, FNP accelerates projection-basis construction by 5–9×. CD uses a small reference set to align the bidirectional image-to-text and text-to-image similarity distributions of the fine-tuned model with CLIP, thereby preserving pretrained cross-modal relations. 
On the 11-task MTIL benchmark, FPCD achieves 87.6\% \emph{Last}, 76.4\% \emph{Average}, and 65.2\% \emph{Transfer}. It also maintains more stable modality gaps and stronger zero-shot retrieval performance on COCO Captions and Flickr30k than competing methods. These results highlight the importance of controlling modality gap drift during continual adaptation of VLMs.
\end{abstract}

\section{Introduction}
\label{sec:intro}



Cross-modal alignment is fundamental to both vision–language models (VLMs) and multimodal large language models (MLLMs). In CLIP-based dual-encoder VLMs~\cite{clip,align,li2021supervision,zhai2023sigmoid}, visual and textual encoders map images and texts into a shared embedding space, where contrastive learning aligns their representations. Such alignment enables strong zero-shot generalization and effective adaptation to downstream tasks. In contrast, MLLMs are typically built upon a large language model and employ a lightweight projector to map visual tokens into the language space~\cite{llava,blip,blip2}, enabling MLLMs to perceive, understand, and reason about visual content.

However, visual and textual modalities differ substantially in information density, semantic granularity, and data structure. Consequently, their representations typically overlap only partially in the shared space, resulting in a modality gap. Recent studies have revealed that this gap is prevalent in multimodal representation spaces and constitutes an intrinsic property of multimodal models~\cite{liang2024foundations, baltruvsaitis2018multimodal,gap,2e1t}. We further verify this phenomenon using CLIP on Flickr30k\cite{flickr30k}. As shown in Figure~\ref{fig:gap}, the measured modality gap is approximately 0.3, indicating that image and text representations still occupy distinct regions of the shared space. Beyond characterizing the separation between modalities, the modality gap also reflects the global geometry of cross-modal alignment. Disrupting this geometry can cause visual and textual representations to drift, thereby degrading performance and generalization.

\begin{figure}[t]
\centering
\includegraphics[width=\linewidth]{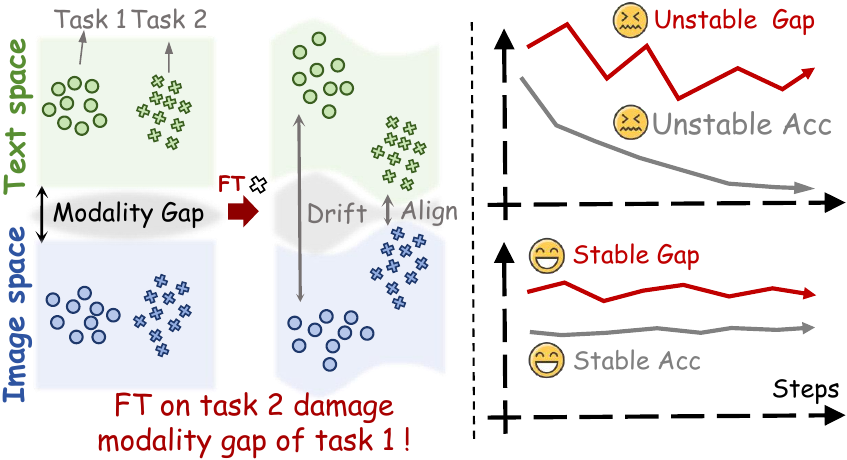}
\caption{Modality gap between Text and Image space. Fine-tuning can damage the gap of other domains. In continual learning, unstable gap can cause performance degradation, thus we hope to maintain a stable gap to preserve the performance.}
\label{fig:demo_gap}
\end{figure}

CLIP is widely adapted to downstream tasks because of its strong generalization ability. However, continual fine-tuning VLMs on a sequence of tasks leads to performance degradation: \textit{catastrophic forgetting} of previously learned knowledge~\cite{McCloskey1989CatastrophicII,Kirkpatrick2016OvercomingCF,Goodfellow2013AnEI} and declining zero-shot performance on unseen tasks. Both are closely related to drift in the pretrained shared embedding space. As illustrated in Fig.~\ref{fig:demo_gap}, fine-tuning a VLM on a new task improves alignment on current task but disrupt it in other domains. Both effects are reflected by changes in modality gap. We argue that the modality gap provides an important indicator of whether the cross-modal geometry is preserved: stable gaps tend to accompany consistent alignment and performance, whereas unintended drift often coincides with degradation. However, most existing continual learning methods primarily focus on downstream task accuracy~\cite{zscl,vpt,pgp,moe_adapter,gift,tang2024mind}. Recent studies have begun to examine the modality gap~\cite{liu2025c,mgclip}, but its relationship with performance remains unclear. This motivates us to study modality-gap dynamics during continual learning.

\begin{figure}[t]
\centering
\includegraphics[width=\linewidth]{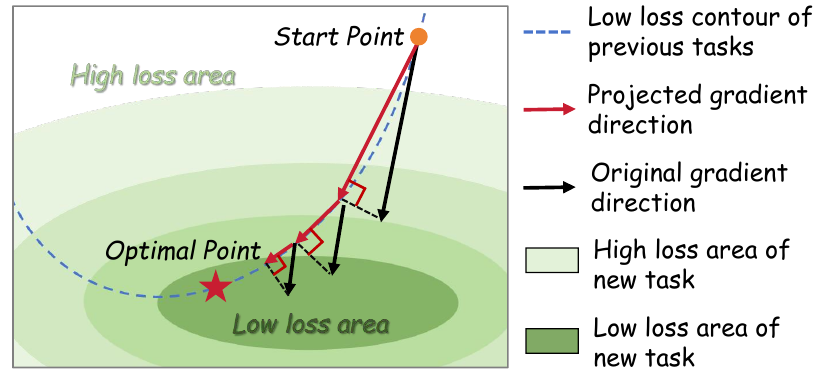}
\caption{Orthogonal Gradient Projection (OGP) projects the gradient of new task (black arrow) onto the null space of previous tasks, using projected gradient (red arrow) to update without compromising the performance on previous tasks.}
\label{fig:demo}
\end{figure}

To address these challenges, we propose \underline{F}ast Null Space \underline{P}rojection and \underline{C}ontrastive \underline{D}istillation CLIP (FPCD), an efficient continual learning method for VLMs that combines Fast Null-space Projection (FNP) and Contrastive Distillation (CD). FNP builds on Orthogonal Gradient Projection (OGP)~\cite{gem,agem,owm,gpm,adam-nscl,dualgpm,yang2025revisiting}, which constrains parameter updates to directions that are orthogonal, or approximately orthogonal, to the feature subspaces of previously tasks, as illustrated in Figure~\ref{fig:demo}. This reduces interference with learned representations and stabilizes the cross-modal embedding geometry. To make gradient projection practical for large-scale VLMs, FNP maintains compact Gram statistics of historical activations instead of storing the complete feature matrices. FNP further exploits the symmetry of the Gram matrices and employs symmetric eigendecomposition rather than conventional singular value decomposition (SVD), accelerating the projection-basis construction by $5$--$9\times$. FNP further adopts a layer-adaptive spectral threshold that balances knowledge preservation and model plasticity according to each layer’s eigenspectrum. Since task-specific activation statistics cannot fully preserve the broad knowledge acquired during CLIP pre-training, we introduce CD to preserve the pretrained cross-modal alignment. Specifically, we use a small reference set and the original CLIP model as a frozen teacher, encouraging the student to match the teacher's bidirectional image-to-text and text-to-image similarity distributions. 
Extensive experiments demonstrate FPCD is effective in mitigating catastrophic forgetting, preserving zero-shot generalization, and maintaining stable cross-modal geometry. 

Our main contributions are:
\begin{itemize}
    \item We systematically analyze modality-gap dynamics during continual learning of VLMs. Results on MTIL and external retrieval datasets show that strong transfer performance does not necessarily imply robust cross-modal generalization, and that unintended gap drift is closely associated with performance degradation.
    \item We propose FNP, an efficient OGP method that accumulates compact Gram statistics and uses symmetric eigendecomposition to accelerate projection-basis construction by 5–9×. A layer-adaptive spectral threshold further balances knowledge preservation and plasticity.
    \item We integrate FNP with contrastive distillation to preserve both previously learned representations and CLIP’s pretrained cross-modal similarity structure. Extensive experiments demonstrate a favorable stability–plasticity–efficiency trade-off across task orders, external retrieval datasets, and model scales.
\end{itemize}

\begin{figure*}[t]
  \centering
  \includegraphics[width=\linewidth]{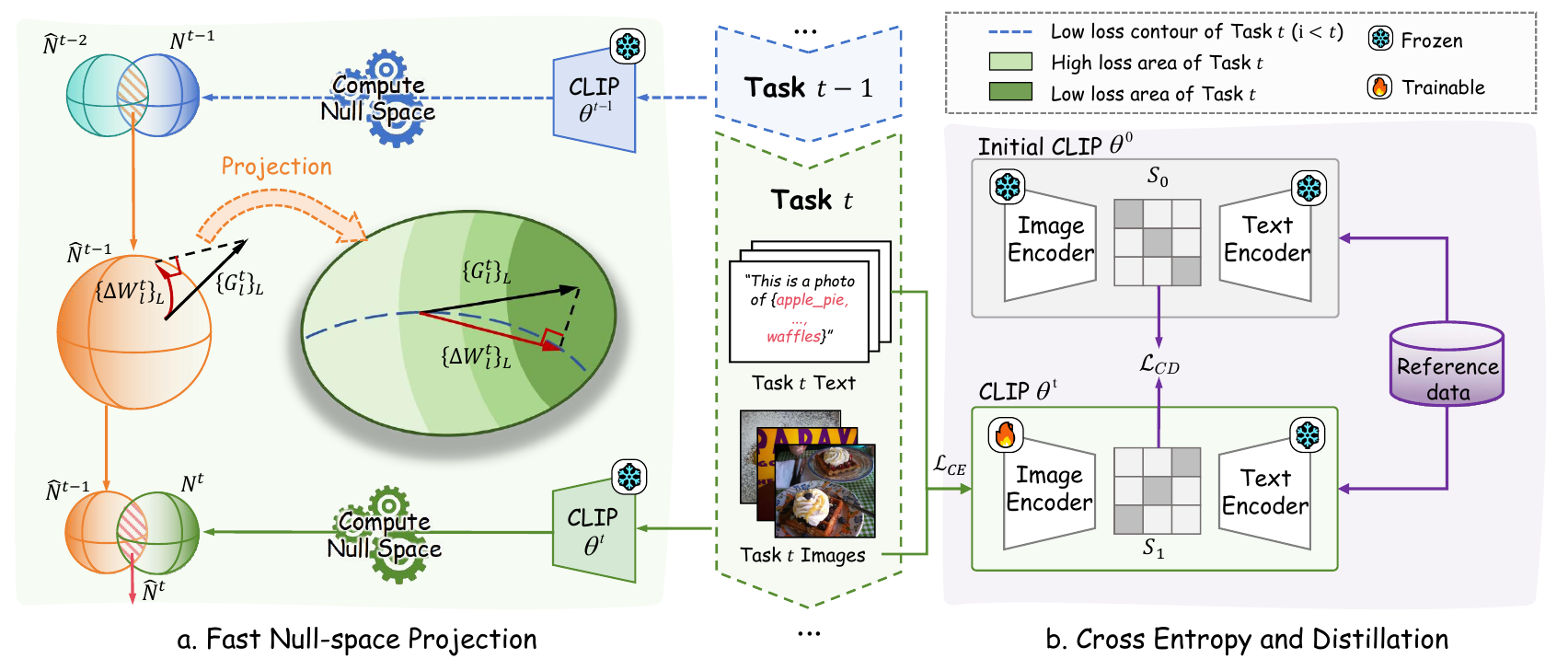}
  \caption{Framework overview of FPCD. (a) \textbf{Fast Null-space Projection}. The updated gradient $G^t_l$ will be projected onto the shared null space $\hat{\mathcal{N}}^{t-1}$ of previous tasks to obtain $\Delta W^t_l$ to update CLIP $\theta^t$. After training task $t$, the null space $\mathcal{N}^{t}$ will be computed to update the shared null space; (b) \textbf{CE Loss and Contrastive Distillation}. Cross Entropy $\mathcal{L}_{CE}$ is used to learning on new task, and Contrastive Distillation $\mathcal{L}_{CD}$ is used for matching CLIP $\theta^t$ and original CLIP $\theta^0$ on reference data.}
  \label{fig:workflow2}
  \vspace{-10pt}
\end{figure*}

\section{Related Work}

\subsection{Continual Learning of Vision–Language Models} 
Pre-trained VLMs are increasingly adapted to sequences of downstream tasks, where the model is expected to acquire task-specific knowledge without forgetting its pre-trained knowledge. Existing approaches can be broadly grouped into three categories.
\textbf{Replay-based methods} revisit stored exemplars, generated samples, or reusable representations from previous tasks~\cite{yan2022generative,lei2023symbolic,zhang2023vqacl,chen2023continual}. Although effective in reducing forgetting, data replay may raise memory and privacy concerns, while generative replay introduces additional computational overhead and potential distribution bias.
\textbf{Cross-modality regularization methods} constrain changes in visual and textual representations through knowledge distillation, alignment preservation, topology constraints, or parameter regularization
~\cite{zscl,zhu2023ctp,cui2024continual,yu2024select,lao2023multi,igsp}. These methods depend strongly on the regularization target, teacher model, and auxiliary data. \textbf{Parameter-efficient adaptation methods} introduce lightweight components while freezing most of the pre-trained backbone, including adapters~\cite{liu2025c,jha2024clap4clip,keeplora,pact}, low‑rank modules~\cite{tang2024mind,lu2024adaptive}, mixture‑of‑experts~\cite{moe_adapter,guo2025hide,forhad2026mixing,xie2026same} and prompts~\cite{wang2022s,d2023multimodal}. While these methods  alleviate forgetting, they rarely examine how continual adaptation alters the cross-modal embedding geometry. Our work complements existing approaches by explicitly analyzing modality-gap dynamics and their relationship with both learned-task retention and zero-shot generalization on external domains.

\subsection{Cross-Modal Geometry and Modality Gap} 
Vision–language models use contrastive learning to align matched image–text pairs while separating unmatched ones in a shared embedding space. The resulting geometry encodes rich semantic structure and supports strong downstream generalization. Nevertheless, image and text representations do not fully overlap; instead, they often occupy partially separated, narrow cone-shaped regions on the hypersphere, a phenomenon known as the modality gap~\cite{gap,2e1t,cone,jiang2023understanding}. Prior studies attribute this gap to factors such as the contrastive objective, nonlinear activations, and imbalanced information content across modalities, suggesting that it is an intrinsic property of multimodal representations. 
During continual fine-tuning, the pretrained embedding geometry may further change through intra-modal displacement and inter-modal rotation, thereby disrupting cross-modal correspondence~\cite{mgclip,modx,NEURIPS2024_71b17f00,mistretta2025cross}. However, most continual learning methods primarily aim to retain task performance and rarely analyze model degradation from the perspective of modality-gap dynamics. The most relevant work, MG-CLIP~\cite{mgclip}, monitors changes in the modality gap to determine when fine-tuning should stop, rather than explicitly constraining its evolution. In this work, we investigate modality gap dynamics during continual fine-tuning and establish a clear empirical association between excessive task-irrelevant gap drift and performance degradation.

\subsection{Orthogonal Gradient Projection for Continual Learning} 
Orthogonal Gradient Projection (OGP) mitigates interference with previously acquired knowledge by constraining updates for a new task to be orthogonal, or approximately orthogonal, to the subspaces of earlier tasks. As illustrated in Fig~\ref{fig:demo}, this mechanism provides an intuitive and interpretable way to balancing knowledge retention and model plasticity. Early studies mainly use OGP to CNNs~\cite{gem,owm,gpm,adam-nscl}, followed by extensions to prompt tuning~\cite{vpt,pgp}, Transformer architectures~\cite{igsp,dynamic}, knowledge editing in LLMs~\cite{alphaedit}, and mixture-of-experts architectures~\cite{xie2026same}. Despite its conceptual advantages, conventional OGP methods remains costly for large-scale VLMs because it often requires storing high-dimensional activation matrices, repeatedly performing expensive singular value decomposition (SVD), and using a fixed projection rank across layers. Such designs incur considerable memory and computational overhead and fail to accommodate the substantially different spectral distributions of individual layers. 
In this work, we develop Fast Null-space Projection, which reduces the storage and decomposition costs of subspace construction and determines a layer-specific approximate null space from each layer’s eigenspectrum.

\section{Method}

\subsection{Preliminaries}
\subsubsection{Continual Learning Setup}
Continual learning is commonly studied under class-incremental learning (CIL) and task-incremental learning (TIL)~\cite{arxiv:TILCIL}. In CIL, new classes are introduced sequentially. In TIL, new domains are introduced sequentially. In this work, we focus on TIL because VLMs are commonly adapted to diverse domains. Compared with CIL, such cross-domain tasks generally exhibit lower task relatedness. This setting reduces the confounding effect of positive transfer across tasks, enabling a cleaner evaluation of whether continual learning preserves the zero-shot generalization.

Specifically, we consider a standard TIL setting in which a model is trained sequentially on a series of $T$ tasks. The $i-th$ task is associated with a dataset
\begin{align}
    \mathcal{D}_i=\left\{\left(I_j^{(i)},T_j^{(i)}\right)\right\}_{j=1}^{N_i},\qquad N_i=\lvert\mathcal{D}_i\rvert
\end{align}
where $I_j^{(i)}$ and $T_j^{(i)}$ denote the image and its class prompt, respectively. When learning task $t$, the model has access only to $D_t$. The objective is to acquire knowledge from the current task while retaining performance on previously learned tasks and preserving zero-shot ability on unseen tasks.

\subsubsection{CLIP-based Vision-Language Models}
Our method is built upon CLIP~\cite{clip}, which embed images and texts into a shared embedding space. The probability of a text description $T_j$ matching an image $I$ is computed by
\begin{align}
    p(T_j|I) = \frac{\exp(\mathrm{cos}(f(I), f(T_j)) / \tau)}{\sum_k \exp(\mathrm{cos}(f(I), f(T_k)) / \tau)},
\end{align}
where $f(\cdot)$ denotes the CLIP model, $\mathrm{cos}(\cdot,\cdot )$ denotes cosine similarity, $\tau$ is a learnable temperature parameter. For fine-tuning on downstream classification, each class is represented by the prompt ``a photo of a \{object\}''. The class with the highest image--text similarity is selected as the prediction, and the model is optimized using the cross-entropy loss $\mathcal{L}_{\mathrm{CE}}$.

\subsubsection{Modality Gap Measurement}

Following~\cite{gap}, we quantify the modality gap between
visual and textual representations using the mean cosine distance
of paired image--text embeddings. Given a dataset
$\mathcal{D}$, we define
\begin{align}
    \mathrm{GAP}(\mathcal{D})
    =
    1-\frac{1}{N}
    \sum_{i=1}^{N}
    \operatorname{cos}(f(I_i),f(T_i)).
    \label{eq:gap}
\end{align}
Larger $\mathrm{GAP}$ indicates greater separation between
two modalities. 

\subsection{Fast Null-space Projection}
We apply FNP to the image encoder, while the same procedure is
applicable to the text encoder. To stabilize the modality
gap, we update only the Feed-Forward Network (FFN) layers, which are
associated with task-specific knowledge~\cite{hase2023does, geva2021transformer, alphaedit}, and freeze the remaining parameters. For task $t$, FNP projects the updates of task $t$ to the common null space of previous task activations, thereby reducing interference with learned representations.


\subsubsection{Formalizing the Null-Space Constraint}
Consider the $l$-th layer of a model with weight matrix $\mathbf{W}_l^{t-1}$ after training on task $t-1$. Let $\mathbf{X}^{t-1}_{l}\in \mathbb{R}^{N_i \times d}$ denote its input activation on data from task $t-1$. The layer output and the input to the subsequent layer are given by:
\begin{align}
    \mathbf{O}_{l}^{t-1}=\mathbf{X}_{l}^{t-1}\mathbf{W}_{l}^{t-1} +\mathbf{b}_{l}^{t-1}, 
    \qquad 
    \mathbf{X}_{l+1}^{t-1}=\sigma\left(\mathbf{O}_{l}^{t-1}\right),
\label{eq:layer_output}
\end{align}
where $\sigma(\cdot)$ is a nonlinear activation function. We freeze the bias term during continual learning, such that $\mathbf{b}_{l}^{t}=\mathbf{b}_{l}^{t-1}$. When learning task $t$, the weight matrix is updated as
\begin{align}
    \mathbf{W}_{l}^{t}=\mathbf{W}_{l}^{t-1}+\Delta\mathbf{W}_{l}^{t}.
\end{align}
Let
$\mathcal{N}_{l}^{t-1}
=\operatorname{Null}(\mathbf{X}_{l}^{t-1})$
denote the null space of the previous task activations. If the
columns of $\Delta\mathbf{W}_{l}^{t}$ lie in
$\mathcal{N}_{l}^{t-1}$, then
\begin{align}
    \mathbf{X}_{l}^{t-1}\Delta\mathbf{W}_{l}^{t}
    =\mathbf{0}, \qquad 
    \mathbf{O}_{l}^{t}
    =
    \mathbf{X}_{l}^{t-1}\mathbf{W}_{l}^{t}
    +\mathbf{b}_{l}^{t}
    =
    \mathbf{O}_{l}^{t-1}.
\label{eq:null_constraint}
\end{align}
Thus, the response of the $l$-th layer to the retained activations
of task $t-1$ remains unchanged after the update. Note that
Eq.~\eqref{eq:null_constraint} just characterizes the effect of the
parameter update at layer $l$, rather than defining an independent
forward process.

To preserve the responses associated with all previous tasks, the
update should lie in their common null space,
\begin{align}
    \widehat{\mathcal{N}}_{l}^{t-1}
    =
    \bigcap_{i=1}^{t-1}\mathcal{N}_{l}^{i}.
\end{align}
Accordingly, we project the gradient of the current task onto
$\widehat{\mathcal{N}}_{l}^{t-1}$. This constraint reduces changes
to previously learned representations and consequently mitigates
unintended drift in the cross-modal embedding space.


\subsubsection{Fast Construction of the Projection Matrix}
Constructing the common null space directly requires storing the
activation matrices $\mathbf{X}_{l}^{i}$ for all previous tasks,
which is impractical for large-scale VLMs. Instead, after learning
task $i$, we compute and store the normalized Gram matrix
\begin{align}
    \mathbf{M}_{l}^{i}
    =
    \frac{
        (\mathbf{X}_{l}^{i})^{\top}\mathbf{X}_{l}^{i}
    }{
        \left\|
        (\mathbf{X}_{l}^{i})^{\top}\mathbf{X}_{l}^{i}
        \right\|_{\mathrm{F}}
    }
    \in\mathbb{R}^{d\times d},
\label{eq:gram}
\end{align}
where $\|\cdot\|_{\mathrm{F}}$ denotes the Frobenius norm. The
historical statistics are accumulated as
\begin{align}
    \widehat{\mathbf{M}}_{l}^{t-1}
    =
    \sum_{i=1}^{t-1}\mathbf{M}_{l}^{i}.
\label{eq:accumulated_gram}
\end{align}
Since each Gram matrix is positive semidefinite, the null space of
their sum equals intersection of their individual null spaces:
\begin{align}
    \operatorname{Null}
    \left(\widehat{\mathbf{M}}_{l}^{t-1}\right)
    =
    \bigcap_{i=1}^{t-1}
    \operatorname{Null}(\mathbf{X}_{l}^{i})
    =
    \widehat{\mathcal{N}}_{l}^{t-1}.
\end{align}
Therefore, the common null space can be recovered without storing
the original activation matrices.

Conventional OGP methods typically apply singular value
decomposition to construct the projection basis. Because
$\widehat{\mathbf{M}}_{l}^{t-1}$ is symmetric and positive
semidefinite, we instead employ symmetric eigendecomposition:
\begin{align}
    \widehat{\mathbf{M}}_{l}^{t-1}
    =
    \mathbf{V}_{l}\boldsymbol{\Lambda}_{l}
    \mathbf{V}_{l}^{\top}, 
    \qquad
    \boldsymbol{\Lambda}_{l}
    =
    \operatorname{diag}
    (\sigma_1,\sigma_2,\ldots,\sigma_d),
\label{eq:eigendecomposition}
\end{align}
where
$\sigma_1\geq\sigma_2\geq\cdots\geq\sigma_d\geq0$.
This replacement accelerates projection-basis construction by
$5$--$9\times$ compared with SVD.

Given an eigenvalue threshold $\sigma_{m,l}$, we partition the
eigenvectors as
\begin{align}
    \mathbf{V}_{l}
    =
    [\mathbf{V}_{l,1},\mathbf{V}_{l,2}],
\end{align}
where $\mathbf{V}_{l,2}$ contains the eigenvectors whose
eigenvalues satisfy $\sigma_j\leq\sigma_{m,l}$. The corresponding
projection matrix is
\begin{align}
    \mathbf{P}_{l}^{t-1}
    =
    \mathbf{V}_{l,2}\mathbf{V}_{l,2}^{\top}.
\label{eq:projection_matrix}
\end{align}
When $\sigma_{m,l}=0$, $\mathbf{V}_{l,2}$ spans the exact null
space. In practice, however, the Gram matrices of CLIP layers
typically exhibit long-tailed eigenspectra with few exactly zero
eigenvalues. We therefore construct an approximate null space by
selecting the largest threshold $\sigma_{m,l}$ satisfying
\begin{align}
    \sum_{\sigma_j\leq\sigma_{m,l}}\sigma_j
    \leq
    \rho\sum_{j=1}^{d}\sigma_j,
\label{eq:adaptive_threshold}
\end{align}
where $\rho$ controls the proportion of spectral energy assigned
to the approximate null space. This layer-adaptive threshold
accounts for layer-wise eigenspectra and
provides a flexible balance between knowledge preservation and
model plasticity.

\subsubsection{Update with Projected Gradients.}
For task $t$, let $\mathbf{G}_{l}^{t}$ denote the original gradient
of the $l$-th layer. We obtain the constrained update direction by
projecting it onto the approximate common null space:
\begin{align}
    \Delta\mathbf{W}_{l}^{t}
    =
    \mathbf{P}_{l}^{t-1}\mathbf{G}_{l}^{t}.
\label{eq:projected_gradient}
\end{align}
Consequently, the projected update $\Delta\mathbf{W}_l^t$ lies in the approximate common null space of the previous task activations while retaining sufficient directions for learning the current task.

FNP stores only one accumulated Gram matrix for each updated
layer. Its memory complexity is $\mathcal{O}(Ld^2)$, where $L$ is the number of updated layers and $d$ is the layer dimension. Importantly, this cost is independent of the number of tasks.

\begin{table*}[t]
\centering
\caption{Comparison of SOTA methods on MTIL Order I. Results of other methods are taken from published papers.}
\begin{tabular}{l|l|ccccccccccc|c}
\toprule
             &Dataset &\rotatebox{70}{Aircraft}&\rotatebox{70}{Caltech101}&\rotatebox{70}{CIFAR100}&\rotatebox{70}{DTD}&\rotatebox{70}{EuroSAT}&\rotatebox{70}{Flowers}&\rotatebox{70}{Food}&\rotatebox{70}{MNIST}&\rotatebox{70}{OxfordPet}&\rotatebox{70}{StanfordCars}&\rotatebox{70}{SUN397}&Avg\\
\midrule
             \multirow[c]{2}{*}{\rotatebox[origin=c]{90}{CLIP}}& 
             Zero-Shot&24.4&87.9&68.2&45.1&54.6&71.3&88.8&59.5&89.1&64.5&64.9&69.4\\
             & Upper Bound&62.0&95.1&89.6&79.5&98.9&97.5&92.7&99.6&94.7&89.6&81.8&89.2\\
\midrule
            \multirow[c]{9}{*}{\rotatebox[origin=c]{90}{Transfer}}
            &LwF~\cite{lwf}
            && 74.5 & 56.9 & 39.1 & 51.1 & 52.6 & 72.8 & 60.6 & 75.1 & 30.3 & 55.9 & 56.89 \\
            &iCaRL~\cite{icarl}
            && 56.6 & 44.6 & 32.7 & 39.3 & 46.6 & 68.0 & 46.0 & 77.4 & 31.9 & 60.5 & 50.36 \\
            &LwF-VR~\cite{lwfvr}
            && 77.1 & 61.0 & 40.5 & 45.3 & 54.4 & 74.6 & 47.9 & 76.7 & 36.3 & 58.6 & 57.24 \\
            &WiSE-FT~\cite{wiseft}
            && 73.5 & 55.6 & 35.6 & 41.5 & 47.0 & 68.3 & 53.9 & 69.3 & 26.8 & 51.9 & 52.34 \\
            &ZSCL~\cite{zscl}
            && 86.0 & 67.4 & 45.4 & 50.4 & 69.1 & 87.6 & 61.8 & 86.8 & 60.1 & 66.8 & 68.14 \\
            &MoE-Adapter\cite{moe_adapter}
            &&87.9&68.2&44.4&49.9&\textbf{70.7}&\textbf{88.7}&59.7&89.1&\textbf{64.5}&65.5&68.86\\
            &DIKI\cite{tang2024mind}
            &&\textbf{92.9}&69.0&43.2&48.2&67.4&85.2&63.0&87.9&63.8&66.2&68.68\\
            &GIFT\cite{gift}
            &&88.5&\textbf{69.8}&46.0&49.4&68.5&87.1&\textbf{69.9}&88.9&57.7&\textbf{67.7}&\textbf{69.35}\\
            \rowcolor{gray!20}&FPCD(Ours)
            &&87.9&67.6&\textbf{46.7}&\textbf{55.0}&70.0&87.9&59.9&\textbf{89.5}&58.9&65.3&68.87\\
\midrule
            \multirow[c]{9}{*}{\rotatebox[origin=c]{90}{Average}}
            &LwF~\cite{lwf}
            & 36.3 & 86.9 & 72.0 & 59.0 & 73.7 & 60.0 & 73.6 & 74.8 & 80.0 & 37.3 & 58.1 & 64.70 \\
            &iCaRL~\cite{icarl}
            & 35.5 & 89.2 & 72.2 & 60.6 & 68.8 & 70.0 & 78.2 & 62.3 & 81.8 & 41.2 & 62.5 & 65.66 \\
            &LwF-VR~\cite{lwfvr}
            & 29.6 & 87.7 & 74.4 & 59.5 & 72.4 & 63.6 & 77.0 & 66.7 & 81.2 & 43.7 & 60.7 & 65.14 \\
            &WiSE-FT~\cite{wiseft}
            & 26.7 & 86.5 & 64.3 & 57.1 & 65.7 & 58.7 & 71.1 & 70.5 & 75.8 & 36.9 & 54.6 & 60.72 \\
            &ZSCL~\cite{zscl}
            & 45.1 & 92.0 & 80.1 & 64.3 & 79.5 & 81.6 & 89.6 & 75.2 & 88.9 & 64.7 & 68.0 & 75.36 \\
            &MoE-Adapter\cite{moe_adapter}
            &50.2&91.9&\textbf{83.1}&69.4&78.9&\textbf{84.0}&89.1&73.7&89.3&\textbf{67.7}&66.9&76.75\\
            &DIKI\cite{tang2024mind}
            &45.1&95.5&\textbf{83.1}&64.8&79.9&83.5&87.0&76.2&89.6&67.0&67.1&76.25\\
            &GIFT\cite{gift}
            &\textbf{51.9}&93.9&81.4&67.7&80.3&82.8&89.3&\textbf{80.6}&90.3&63.1&\textbf{68.9}&77.29\\
            \rowcolor{gray!20}&FPCD(Ours)
            &51.7&\textbf{95.5}&82.6&\textbf{69.3}&\textbf{82.7}&82.9&\textbf{90.0}&74.3&\textbf{91.0}&63.7&66.7&\textbf{77.31}\\
\midrule
            \multirow[c]{10}{*}{\rotatebox[origin=c]{90}{Last}}
            &LwF~\cite{lwf}
            & 26.3 & 87.5 & 71.9 & 66.6 & 79.9 & 66.9 & 83.8 & 99.6 & 92.1 & 66.1 & 80.4 & 74.65 \\
            &iCaRL~\cite{icarl}
            & 35.8 & 93.0 & 77.0 & 70.2 & 83.3 & 88.5 & 90.4 & 86.7 & 93.2 & 81.2 & \textbf{81.9} & 80.11 \\
            &LwF-VR~\cite{lwfvr}
            & 20.5 & 89.8 & 72.3 & 67.6 & 85.5 & 73.8 & 85.7 & 99.6 & 93.1 & 73.3 & 80.9 & 76.55 \\
            &WiSE-FT~\cite{wiseft}
            & 27.2 & 90.8 & 68.0 & 68.9 & 86.9 & 74.0 & 87.6 & 99.6 & 92.6 & 77.8 & 81.3 & 77.70 \\
            &ZSCL~\cite{zscl}
            & 40.6 & 92.2 & 81.3 & 70.5 & 94.8 & 90.5 & 91.9 & 98.7 & 93.9 & 85.3 & 80.2 & 83.63 \\
            &MoE-Adapter\cite{moe_adapter}
            &\textbf{49.8}&92.2&86.1&\textbf{78.1}&95.7&94.3&89.5&98.1&89.9&81.6&80.0&85.03\\
            &DIKI\cite{tang2024mind}
            &45.2&95.7&\textbf{86.3}&72.9&98.0&\textbf{97.0}&89.2&\textbf{99.4}&94.2&81.6&76.6&85.10\\
            &GIFT\cite{gift}
            &47.9&95.6&82.8&75.1&97.3&94.2&91.7&99.2&94.2&87.0&80.9&85.99\\
            \rowcolor{gray!20}&FPCD(Ours)
            &46.4&\textbf{95.9}&85.1&77.0&\textbf{98.2}&93.3&\textbf{92.7}&99.3&\textbf{94.9}&\textbf{87.6}&80.5&\textbf{86.45}\\
\midrule
\bottomrule
\end{tabular}
\label{tab:mtil_1}
\end{table*}

\subsection{Contrastive Distillation}
FNP preserves representations associated with previous tasks, but
cannot protect the broader cross-modal geometry learned during CLIP
pre-training because the pre-training data are unavailable. We
therefore introduce Contrastive Distillation (CD)~\cite{kd,gift},
using the original CLIP as a frozen teacher and 1,000 ImageNet
data~\cite{imagenet} as a reference set.

For a reference batch $\{(I_i,T_i)\}_{i=1}^{B}$, let
$m\in\{0,t\}$ denote the teacher and the current model, respectively.
Their cross-modal similarity matrices are defined as
\begin{align}
    S_{i,j}^{m}
    &=
    \operatorname{cos}
    \left(f_I^{m}(I_i),f_T^{m}(T_j)\right),
    \qquad
    \mathbf{Z}^{m}=\frac{\mathbf{S}^{m}}{\tau_{\mathrm{d}}},
\label{eq:cd_similarity}
\end{align}
where $\tau_{\mathrm{d}}$ is the distillation temperature. CD matches
the teacher and student similarity distributions in both directions:
\begin{align}
    \mathcal{L}_{\mathrm{CD}}
    &=
    \frac{1}{B}\sum_{i=1}^{B}
    D_{\mathrm{KL}}\!\left(
    \operatorname{softmax}(\mathbf{Z}^{0}_{i,:})
    \,\Vert\,
    \operatorname{softmax}(\mathbf{Z}^{t}_{i,:})
    \right)
    \nonumber\\
    &\quad+
    \frac{1}{B}\sum_{j=1}^{B}
    D_{\mathrm{KL}}\!\left(
    \operatorname{softmax}(\mathbf{Z}^{0}_{:,j})
    \,\Vert\,
    \operatorname{softmax}(\mathbf{Z}^{t}_{:,j})
    \right)
\label{eq:cd_loss}
\end{align}
The two terms preserve the image-to-text and text-to-image
similarity structures, respectively.

\subsection{Overall Training Objective}
For each task, the model is optimized using the combination of the classification objective and contrastive distillation:
\begin{equation}
    \mathcal{L} = \mathcal{L}_{\text{CE}} + \mathcal{L}_{\text{CD}}. 
\end{equation}
Here, $\mathcal{L}_{\text{CE}}$ adapts the model to the current task, while $\mathcal{L}_{\mathrm{CD}}$ preserves the bidirectional cross-modal similarity distributions of the original CLIP model on the reference data. As shown in Figure~\ref{fig:workflow2}, FNP projects the gradient of the overall objective before each parameter update $\Delta \mathbf{W}^{t}_{l}=\mathbf{P}^{t-1}_{l}\mathbf{G}^{t}_{l}$. It reduces changes to feature responses associated with previously learned tasks, whereas CD preserves the broader pretrained cross-modal relations. Their combination stabilizes both learned-task features and CLIP's cross-modal geometry, thereby mitigating catastrophic forgetting and unintended modality-gap drift while retaining zero-shot generalization.


\begin{table*}[t]
\centering
\caption{Comparison of SOTA methods on MTIL Order II. Results of other methods are taken from published papers.}
\begin{tabular}{l|l|ccccccccccc|c}
\toprule
             &Dataset &\rotatebox{70}{StanfordCars}&\rotatebox{70}{Food}&\rotatebox{70}{MNIST}&\rotatebox{70}{OxfordPet}&\rotatebox{70}{Flowers}&\rotatebox{70}{SUN397}&\rotatebox{70}{Aircraft}&\rotatebox{70}{Caltech101}&\rotatebox{70}{DTD}&\rotatebox{70}{EuroSAT}&\rotatebox{70}{CIFAR100}&Avg\\
\midrule
             \multirow[c]{2}{*}{\rotatebox[origin=c]{90}{CLIP}}& 
             Zero-Shot&64.5&88.8&59.5&89.1&71.3&64.9&24.4&87.9&45.1&54.6&68.2&65.4\\
             & Upper Bound&89.6&92.7&99.6&94.7&97.5&81.8&62.0&95.1&79.5&98.9&89.6&89.2\\
\midrule
            \multirow[c]{9}{*}{\rotatebox[origin=c]{90}{Transfer}}
            &LwF~\cite{lwf}
            && 87.8 & 58.5 & 71.9 & 46.6 & 57.3 & 12.8 & 81.4 & 34.5 & 34.5 & 46.8 & 53.21 \\
            &iCaRL~\cite{icarl}
            &&86.1 & 51.8 & 67.6 & 50.4 & 57.9 & 11.0 & 72.3 & 31.2 & 32.7 & 48.1 & 50.91 \\
            &LwF-VR~\cite{lwfvr}
            && 88.2 & 57.0 & 71.4 & 50.0 & 58.0 & 13.0 & 82.0 & 34.4 & 29.3 & 47.6 & 53.09 \\
            &WiSE-FT~\cite{wiseft}
            && 87.2 & 57.6 & 67.0 & 45.0 & 54.0 & 12.9 & 78.6 & 35.5 & 28.4 & 44.3 & 51.05 \\
            &ZSCL~\cite{zscl}
            && 88.3 & 57.5 & 84.7 & 68.1 & 64.8 & 21.1 & 88.2 & 45.3 & \textbf{55.2} & 68.2 & 64.14 \\

            &MoE-Adapter\cite{moe_adapter}
            &&\textbf{88.8}&59.5&89.1&69.9&64.4&18.1&86.9&43.7&54.6&68.2&64.32\\
            &DIKI\cite{tang2024mind}
            &&85.8&59.8&89.1&\textbf{71.8}&62.6&\textbf{24.3}&\textbf{93.3}&42.7&46.8&67.8&64.40\\
            &GIFT\cite{gift}
            &&88.3&\textbf{63.4}&88.1&70.8&\textbf{67.7}&22.8&90.4&46.7&51.8&68.8&\textbf{65.88}\\
            
            \rowcolor{gray!20}&FPCD(Ours)
            &&88.5&59.5&\textbf{89.2}&70.7&66.1&21.2&88.0&\textbf{47.3}&52.9&\textbf{69.1}&65.25\\
\midrule
            \multirow[c]{9}{*}{\rotatebox[origin=c]{90}{Average}}
            &LwF~\cite{lwf}
            & 49.0 & 77.0 & 92.1 & 85.9 & 66.5 & 67.2 & 20.9 & 84.7 & 44.6 & 45.5 & 50.5 & 62.17 \\
            &iCaRL~\cite{icarl}
            & 52.0 & 75.9 & 77.4 & 74.6 & 58.4 & 59.3 & 11.7 & 79.6 & 42.1 & 43.2 & 51.7 & 56.90 \\
            &LwF-VR~\cite{lwfvr}
            & 44.9 & 75.8 & 91.8 & 85.3 & 63.5 & 67.6 & 16.9 & 84.9 & 44.0 & 40.6 & 51.3 & 60.60 \\
            &WiSE-FT~\cite{wiseft}
            & 52.6 & 79.3 & 91.9 & 83.9 & 63.4 & 65.2 & 23.3 & 83.7 & 45.4 & 40.0 & 48.2 & 61.54 \\
            &ZSCL~\cite{zscl}
            & 81.7 & 91.3 & 91.9 & 91.0 & 82.9 & 72.5 & 33.6 & 89.7 & 53.3 & \textbf{62.8} & 69.9 & 74.60 \\

            &MoE-Adapter\cite{moe_adapter}
            &84.9&89.9&89.3&91.4&86.2&72.2&33.4&89.4&53.3&61.4&69.9&74.66\\
            &DIKI\cite{tang2024mind}
            &81.9&88.9&92.1&92.8&\textbf{87.7}&70.3&34.3&94.2&51.5&56.1&69.5&74.48\\
            &GIFT\cite{gift}
            &83.2&90.8&\textbf{92.6}&92.8&85.8&\textbf{74.1}&36.0&\textbf{92.1}&54.7&60.0&70.4&75.68\\
            
            
            \rowcolor{gray!20}&FPCD(Ours)
            &\textbf{88.1}&\textbf{91.6}&92.0&\textbf{93.1}&86.4&73.4&\textbf{36.7}&91.0&\textbf{56.2}&61.0&\textbf{70.7}&\textbf{76.38}\\
\midrule
            \multirow[c]{9}{*}{\rotatebox[origin=c]{90}{Last}}
            &LwF~\cite{lwf}
            & 34.6 & 69.6 & 99.3 & 88.7 & 61.1 & 72.5 & 32.5 & 88.1 & 65.6 & 90.9 & 87.9 & 71.89 \\

            &iCaRL~\cite{icarl}
            & 46.0 & 81.5 & 91.3 & 82.8 & 66.5 & 72.2 & 16.3 & 91.6 & 68.1 & 83.2 & 87.8 & 71.57 \\

            &LwF-VR~\cite{lwfvr}
            & 27.4 & 61.2 & 99.4 & 86.3 & 60.6 & 70.7 & 23.4 & 88.0 & 61.3 & 84.3 & \textbf{88.1} & 68.25 \\

            &WiSE-FT~\cite{wiseft}
            & 35.6 & 76.9 & \textbf{99.5} & 89.1 & 62.1 & 71.8 & 27.8 & 90.8 & 67.0 & 85.6 & 87.6 & 72.16 \\

            &ZSCL~\cite{zscl}
            & 78.2 & 91.1 & 97.6 & 92.5 & 87.4 & 78.2 & 45.0 & 92.3 & 72.7 & 96.2 & 86.3 & 83.41 \\

            &MoE-Adapter\cite{moe_adapter}
            &84.1&88.5&94.0&91.8&94.1&77.8&50.4&93.3&77.1&87.7&86.6&84.13\\
            &DIKI\cite{tang2024mind}
            &81.9&89.2&99.4&94.3&\textbf{96.8}&76.7&46.3&95.9&74.8&\textbf{98.3}&86.6&85.47\\
            &GIFT\cite{gift}
            &81.0&90.2&98.6&94.0&91.5&\textbf{78.6}&51.7&94.6&75.6&95.4&86.6&85.25\\
            
            \rowcolor{gray!20}&FPCD(Ours)
            &\textbf{87.4}&\textbf{91.6}&99.3&\textbf{94.5}&95.4&78.1&\textbf{56.5}&\textbf{96.2}&\textbf{79.3}&97.9&87.0&\textbf{87.56}\\
\midrule
\bottomrule
\end{tabular}
\label{tab:mtil_2}
\end{table*}



\section{Experiments}

\subsection{Experiments Setup} 

\subsubsection{Model and Benchmarks}
Main experiments are conducted on CLIP ViT-B/16. We evaluate our method primarily on the MTIL~\cite{zscl}, a widely used benchmark for VLMs that consists of 11 datasets from diverse domains. We also conducted Image-to-Text retrieval task on COCO~\cite{cococaptions} and Flickr30k~\cite{flickr30k} to evaluate the cross-modal zero-shot capability.

\subsubsection{Evaluation Metrics} 
For MTIL, we report three metrics~\cite{zscl}:
a) \emph{Last}, which measures the average accuracy
over all tasks after completing the entire training; 
b) \emph{Transfer}, which evaluates zero-shot generalization on unseen tasks; and
c) \emph{Average}, which summarizes the overall performance throughout the continual learning process.
For image-to-text retrieval on COCO Captions and Flickr30K, we
report Recall@1.

\subsubsection{Implementation Details}
We fine-tune the 12 Feed-Forward Network (FFN) layers within the image encoder of CLIP ViT-B/16 model, while keeping other parameters frozen. We use learning rate of $1\times 10^{-6}$, iteration of 1000 and use a batch size of 64, and eigenvalue ratio for FNP is $\rho=0.15$.

\begin{figure*}[t]
    \centering
    \includegraphics[width=0.9\linewidth]{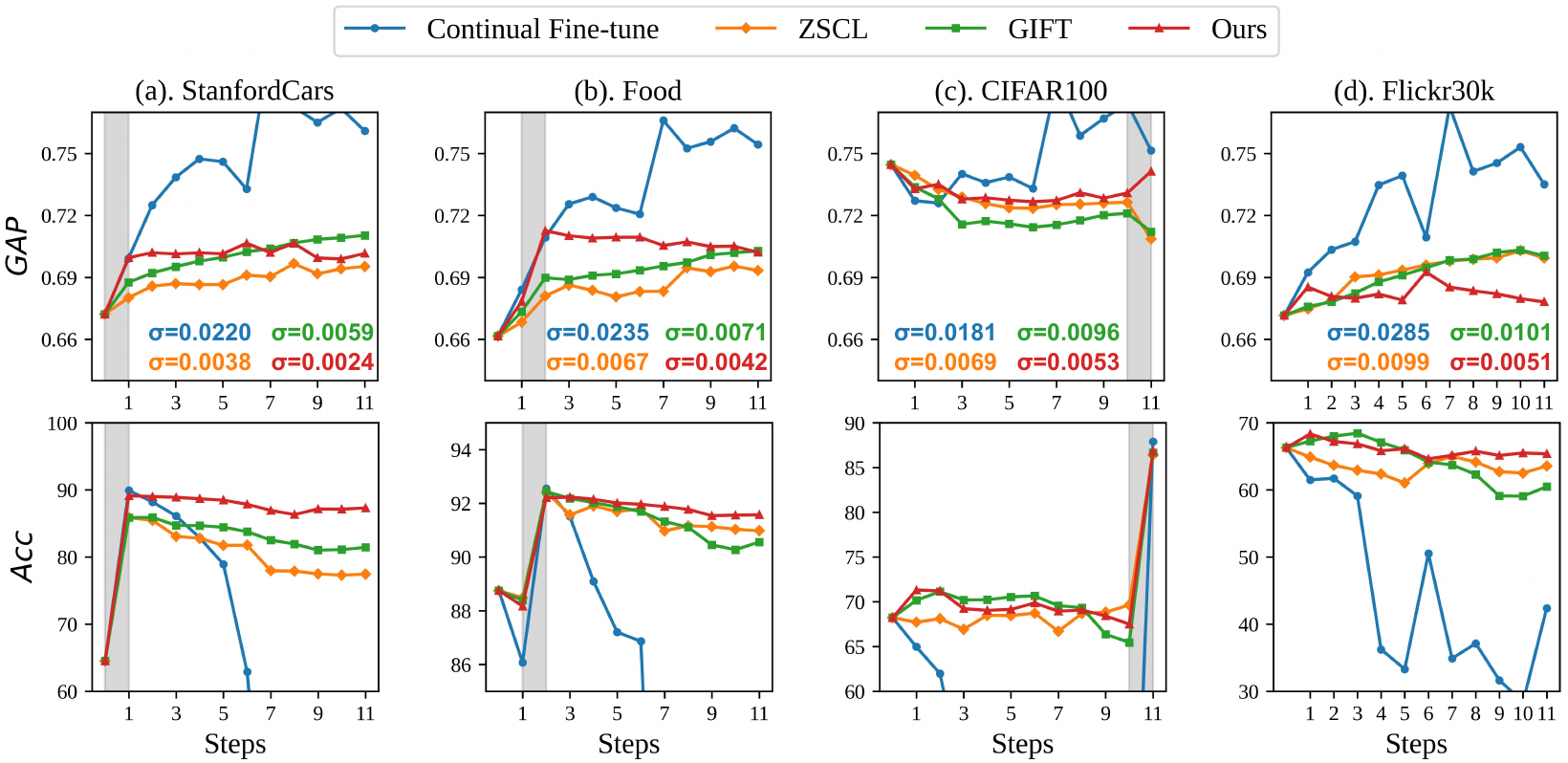}
    \caption{\textbf{Upper:} The variation of modality gap when fine-tuning using different methods, with the gray area representing the fine-tuning position. All have undergone significant changes before and after fine-tuning. \textbf{Bottom:} The variation of accuracy and recall@1. Our method achieves the most stable modality gap and the best performance.}
    \label{fig:gap}
\end{figure*}

\subsection{Main Results on MTIL}
Table~\ref{tab:mtil_1} and Table~\ref{tab:mtil_2} compare FPCD with state-of-the-art continual learning methods on the MTIL benchmark. Zero-shot denotes the performance of the original CLIP without fine-tuning, whereas Upper Bound reports the task-specific performance obtained by independent fine-tuning. Overall, FPCD achieved competitive performance under both task orders, demonstrating a strong balance between plasticity, stability and generalization. 
Specifically, 
1) Excellent fine-tuning performance: FPCD achieves the highest \emph{Last} under both task orders, indicating strong adaptation to downstream tasks while effectively retaining previously acquired knowledge.. 
2) Generalization Preservation:  Its \emph{Transfer} peformance remains close to that of Zero-shot CLIP and is competitive with the best-performing methods. This confirms that our modality gap preservation strategies and knowledge distillation successfully preserve generalizable knowledge inherited from pre-training; 
3) Overall Performance: FPCD achieves the highest \emph{Average} on both order, further demonstrating its strong overall performance throughout the continual learning process.

We observe anomalously high \emph{Transfer} performance on MNIST
for some methods. For example, GIFT~\cite{gift} achieves 69.9\%,
substantially exceeding the 59.5\% of original CLIP,
whereas others remain close to the latter. This
atypical gain disproportionately increases GIFT's overall
\emph{Transfer} score and may overstate its ability to
preserve generalization. To reduce the influence, we adopt Order II as the primary setting for subsequent analyses, in which MNIST is introduced earlier.


\subsection{Analysis on Modality Gap}
While the previous section demonstrates the accuracy advantages of FPCD, it remains important to understand how continual fine-tuning affects the structure of shared embedding space. We investigate the dynamics of the modality gap during continual learning and its relationship with downstream performance. Following Eq.~\ref{eq:gap}, the gap is estimated using 200 randomly sampled image–text pairs from each MTIL test set and 2,000 pairs from Flickr30k~\cite{flickr30k}. Let $GAP(\mathcal{D}_i)_k$ denote the gap of dataset $\mathcal{D}_i$ after training step $k$, where $GAP(\mathcal{D}_i)_0$ corresponds to the original CLIP model.

\subsubsection{Fine-tuned tasks exhibit intentional gap changes}
The gray regions in Figure~\ref{fig:gap}indicate the training step at which each dataset is directly fine-tuned. Every time a dataset is fine-tuned, its own modality gap undergoes a sharp changes.
This behavior reflects that the model is strengthening cross-modal alignment for that specific task, as fine-tuning optimizes cosine similarity between its image and text representations. Such changes are expected and beneficial for the current task. However, the key question is whether other tasks experience unintended distortions.

\subsubsection{Other tasks suffer gap drift}
Outside the fine-tuning steps (non-gray regions), Figure~\ref{fig:gap} reveals substantial differences among baselines. Continual Fine-tuning shows large oscillations in $GAP(\mathcal{D}_i)$, indicating uncontrolled changes to the embedding space. ZSCL and GIFT also show a continuous changing gap, meaning that their updates collapse the alignment between modalities across unrelated domains. Such findings demonstrate that existing CL strategies unintentionally rotate or distort representations of unseen or previously learned tasks, harming the global structure of shared embedding space.

\subsubsection{Gap stability correlates with performance}
\begin{figure}[h]
\centering
\includegraphics[width=0.9\linewidth]{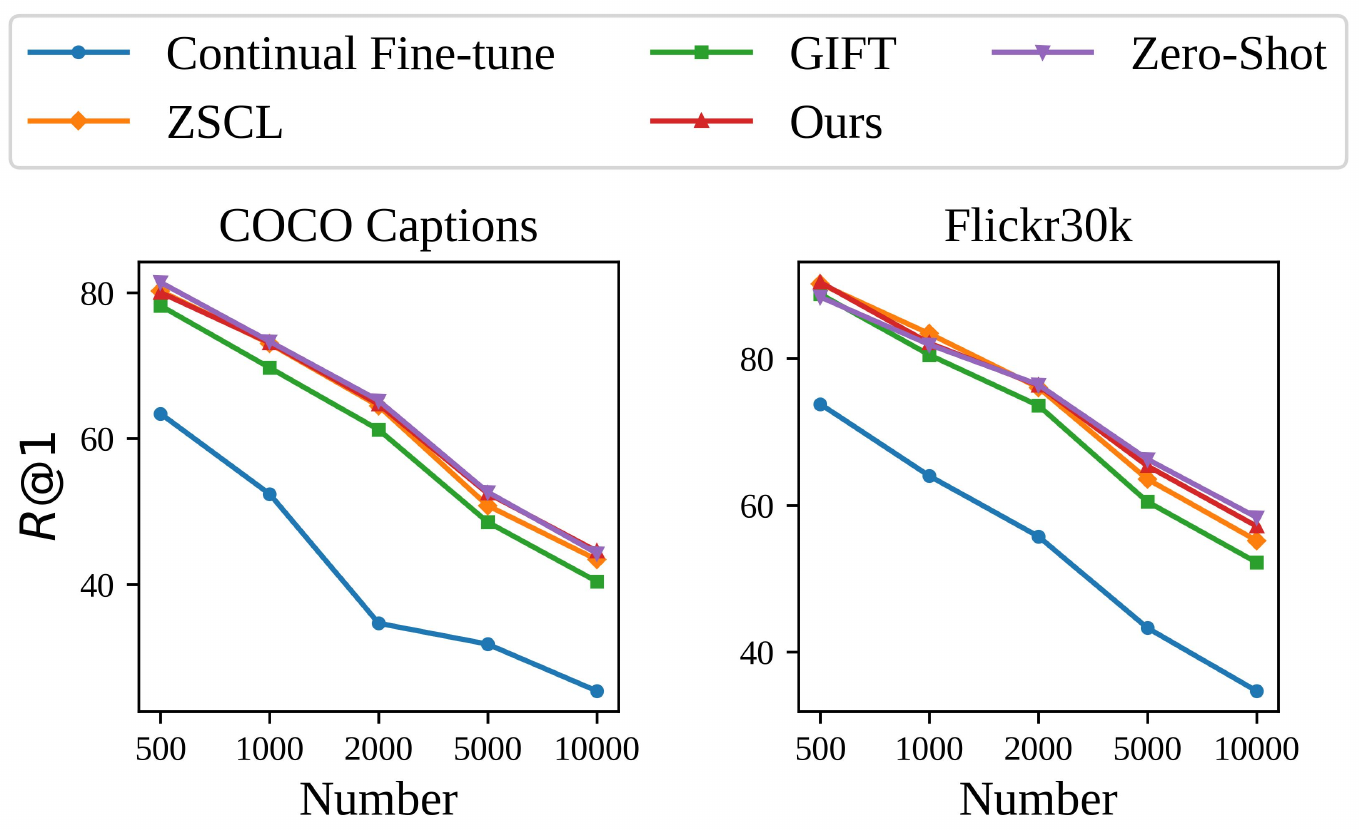}
\caption{Comparison of retrieval task on COCO Captions 
and Flickr30k after training on MTIL. FPCD is closest to original CLIP. The horizontal axis represents the number of data used}
\label{fig:i2t}
\end{figure}

To quantitatively measure the stability of modality gap, we compute the standard deviation $\sigma$ of $\widehat{\mathrm{GAP}}(\mathcal{D}_i)_k$ sequence after removing the discontinuity at step $i$

\begin{equation}
\widehat{\mathrm{GAP}}(\mathcal{D}_i)_k
=
\begin{cases}
\mathrm{GAP}(\mathcal{D}_i)_k,
& k < i, \\[2pt]
\mathrm{GAP}(\mathcal{D}_i)_k
+ \Delta_{FT}\mathrm{GAP}(\mathcal{D}_i),
& k > i,
\end{cases}
\label{eq:adjusted_gap}
\end{equation}
where $\Delta_{FT}\mathrm{GAP}(\mathcal{D}_i) =\mathrm{GAP}(\mathcal{D}_i)_{i-1}-\mathrm{GAP}(\mathcal{D}_i)_i$, and $k\in[0,10],k\neq i$. As reported on Figure~\ref{fig:gap} upper, FPCD achieves the lowest gap variation on all four datasets. 
The bottom shows the corresponding classification accuracy and Recall@1 on Flickr30k, for methods with unstable modality gaps (large $\sigma$), performance sharply drops after the fine-tuning steps. ZSCL and GIFT, even though they achieve relatively strong \emph{Transfer} on MTIL, show clear declines on Flickr30k, demonstrating degradation of zero-shot generalization. This strong correlation between modality gap stability and downstream performance suggests that the modality gap captures the integrity of cross-modal alignment, and its drift provides a reliable proxy for embedding-space collapse. Across all datasets, FPCD shows the smallest gap variance (lowest $\sigma$), meaning its updates induce minimal unintended distortion to the embedding space, and this stability translates into more reliable performance, achieves the highest and most stable accuracy across tasks in MTIL and Image-to-Text retrieval. Figure~\ref{fig:i2t} shows $R@1$ on COCO Captions~\cite{cococaptions} and Flickr30k~\cite{flickr30k}. ZSCL, GIFT both showed decline, while FPCD consistently close to CLIP.


\subsection{Ablation Studies}
Having validated the importance of modality gap stability, we now analyze how each component of FPCD and the impact of key hyper-parameters contribute to this behavior. 

\subsubsection{Component Analysis.}
We progressively added each component to a standard continual fine-tune baseline, and results are presented in Table~\ref{tab:ablation}.
\textbf{FNP} alone significantly improves the \emph{Last} metric, and has already surpassed the previous SOTA method (ours 86.5\% vs DIKI 85.5\%) on \emph{Last}, demonstrating excellent anti-forgetting ability. Additionally, due to its limitation of a low rank null space when updating parameters, it also retains generalization ability to some extent, with \emph{Transfer} reaching 61.3\%. \textbf{CD} performs well in maintaining \emph{Transfer} performance by distilling CLIP knowledge, but does not limit forgetting of learned tasks, resulting in lower \emph{Last}. The combination of FNP and CD has generated a strong synergistic effect, surpassing the previous SOTA method in both \emph{Avg} and \emph{Last} while maintaining high \emph{Transfer} performance. In addition, we also compared CD with feature-level distillation FD. It can be observed that the FD's strict feature constraints greatly limit the model's ability, significantly weaker than CD.

\begin{table}[t]
\centering
\caption{Ablation study of FNP, CD, and FD.}
\begin{tabular}{ccc|ccc}
\toprule
\multicolumn{3}{c}{Method} &  \emph{Transfer} &  \emph{Avg}  &  \emph{Last}\\ 
\midrule
\multicolumn{3}{l|}{Zero-shot} &  65.4  &  65.3  &  65.3 \\ 
\midrule
+FNP   & +CD   & +FD &  \emph{Transfer} &  \emph{Avg}  &  \emph{Last}  \\ 
\midrule
  & & {\ding{51}} & 58.8 & 68.5 & 76.9  \\
 & {\ding{51}} & & 64.0 & 72.7 & 80.7  \\
{\ding{51}} &   & & 61.3 & 73.4 & 86.5 \\
{\ding{51}} &   & {\ding{51}}& 61.5 & 73.7 & 87.1 \\
\rowcolor{gray!20} {\ding{51}} & {\ding{51}} & & \textbf{65.2} & \textbf{76.4} & \textbf{87.6} \\
\bottomrule
\end{tabular}

\label{tab:ablation}
\end{table}

\begin{table}[t]
\centering
\caption{Ablation study on hyperparameter choices. Default settings are marked by underline.}
\centering
\small
\begin{tabular}{ll|ccc}
    \toprule
    Category & Setting & \multicolumn{1}{c}{\emph{Transfer}} & \multicolumn{1}{c}{\emph{Avg}} & \multicolumn{1}{c}{\emph{Last}} \\
    \midrule
    \multirow{5}{*}{Projection} 
    & $\rho=0.1$ & \textbf{65.3} & 76.2 & 86.8 \\
    & \underline{$\rho=0.15$} & 65.2 & \textbf{76.4} & \textbf{87.6} \\ 
    & $\rho=0.2$ & 65.1 & 76.3 & 87.3 \\
    &Dimension = $d/2$ &65.2&76.0&86.9\\
    &$\sigma_m=50\sigma_{min}$~\cite{adam-nscl} &64.7&73.9&83.6\\
    \midrule
    \multirow{3}{*}{\makecell{Ref. Data}} 
    &\underline{ImageNet} & \textbf{65.2} & \textbf{76.4} & \textbf{87.6} \\
    & current & 64.2 & 73.6 & 83.1 \\
    &Syn(1k)\cite{gift} & 64.0 & 75.0 & 86.3 \\
    \midrule
    \multirow{3}{*}{Teacher}
    & \underline{CLIP} & \textbf{65.2} & \textbf{76.4} & \textbf{87.6} \\
    & Last & 64.9 & 76.1 & 87.2 \\
    & WISE(0.5) & 65.0 & 76.1 & 87.1 \\
    \bottomrule
\end{tabular}

\label{tab:hyperp}
\end{table}

\subsubsection{Distillation Design}
The selection of reference data and teacher models is a key factor in Contrastive Distillation, as shown in Table~\ref{tab:hyperp}. In terms of data sources and teacher models, the Synthetic(1k) refers to 1k images synthesized from Stable Diffusion like GIFT~\cite{gift}, the current refers to the current training data, the Last model refers to the latest model trained on the previous dataset, and the WISE(0.5) refers to the model obtained using the WISE\cite{wiseft} method with a coefficient of 0.5. The combination of ImageNet and CLIP achieved the best results, especially in the \emph{Transfer} metric. It means this combination can better preserve the generalization ability of the model. Compared to previous methods (ZSCL~\cite{zscl} requires 1M ImageNet images, GIFT~\cite{gift} relies on diffusion model), our method only requires 1k images from ImageNet, resulting in lower acquisition costs. 

\subsubsection{Null-Space Dimension.}
\begin{figure}[t]
\centering
\includegraphics[width=0.9\linewidth]{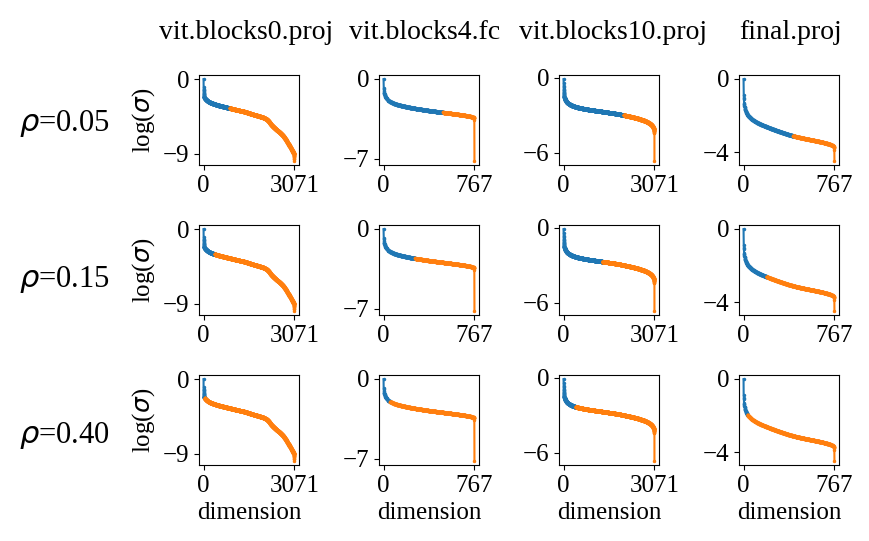}
\caption{Eigenvalues of gram matrix at different FFN layers of CLIP. The eigenvalues smaller than $\sigma_m$ are colored in orange. As $\rho$ increases, the dimension of null space increases.}
\label{fig:singular_value}
\end{figure}

The eigenvalue ratio $\rho$ controls the threshold $\sigma_m$, which determines the rank of the projection matrix. Figure~\ref{fig:singular_value} shows the eigenvalue spectra of the Gram matrices $M_l^{0}$ from several FFN layers, where eigenvalues below $\sigma_m$ are highlighted in orange. A larger $\rho$ generally increases $\sigma_m$, producing a higher-dimensional approximate null space and allowing updates along more directions, which can enhance the model's plasticity but also cause forgetting. As shown in Table~\ref{tab:hyperp}, our layer-adaptive thresholding strategy consistently outperforms the fixed-dimensional and fixed-threshold alternatives used in \cite{adam-nscl,alphaedit}. Increasing $\rho$ from $0.1$ to $0.15$ improves the overall stability--plasticity trade-off, whereas a further increase to $0.2$ slightly degrades both \emph{Transfer} and \emph{Last}, indicating increased forgetting. The best balance is achieved at $\rho=0.15$.

\begin{table}[t]
\centering
\caption{Comparison of computational cost on MTIL in terms of training parameters, GPU burdens and training speed. The $\Delta$ is the improvement relative to GIFT. Results marked with $\dagger$ are reproduced by us; others are taken from paper ~\cite{moe_adapter}.}
\begin{tabular}{l|ccc}
\toprule
 \multicolumn{1}{c}{Method} & Params & GPU Memory & Time (s/iter) \\ 
\midrule
LWF\cite{lwf}           &  149.6M      &  31.4GB  &  1.54s/it \\
ZSCL\cite{zscl}  &  149.6M      &  25.7GB  &  3.94s/it \\
Adapters\cite{moe_adapter} &  51.1M      &  21.8GB  &  1.58s/it \\
GIFT$\dagger$ \cite{gift}    &  149.6M &  34.0GB & 2.37s/it \\
\midrule
FNP-only & 57M & 10.8GB & 0.63s/it \\
FNP+CD & 57M & 19.6GB & 1.48s/it \\
$\Delta$ to GIFT & {\textbf{-61.9\%}} & {\textbf{-42.4\%}} & {\textbf{-37.6\%}} \\
\bottomrule
\end{tabular}

\label{tab:computing}
\end{table}

\subsection{Computational Efficiency}
Beyond accuracy improvements, we also evaluate the computational efficiency of FPCD. All experiments are conducted on NVIDIA RTX A6000 GPU with 48GB Memory. Table~\ref{tab:computing} shows that our method is superior to the SOTA method GIFT~\cite{gift}, with a reduction of approximately 62\%, 42\%, and 38\% in terms of training parameter (M), GPU burdens (GiB), and iteration time, respectively. We further analyze the efficiency of FNP and CD. FNP-only has the smallest GPU memory consumption and the fastest training speed, proving projection operation during the training only brings a small consumption. While the distillation technique, brings more GPU memory consumption and reduces training speed. Besides, The FNP only requires 462MiB to store gram matrix, and calculation of projection matrix is completed before the start of each task, only takes within seconds.
\label{sec:exp:Computing Resources}




\section{Conclusion}
\label{sec:conclusion}

In this work, we investigated how continual fine-tuning affects the shared embedding geometry of VLMs. Our analysis shows that sequential adaptation induces unintended modality gap, which is closely associated with degradation in learned-task performance and zero-shot generalization. We proposed FPCD, which combines Fast Null-space Projection (FNP) and Contrastive Distillation (CD). FNP constrains parameter updates using layer-adaptive projection subspaces constructed from compact historical Gram statistics, while CD preserves CLIP’s pretrained cross-modal similarity structure through bidirectional distribution matching. Experiments demonstrate that FPCD effectively balances knowledge retention, continual adaptation, and zero-shot generalization while maintaining a stable modality gap. These results highlight the importance of controlling unintended cross-modal geometry drift during continual adaptation of VLMs.
\section{Limitations and Future Work}
\label{sec:limitations}
Despite its favorable stability–plasticity trade-off, FPCD has several limitations. First, FNP use accumulated statistics, which may blur task-specific directions and reduce the subspace available for learning new knowledge. More compact subspace representations may improve scalability.

Second, FNP provides an approximate projection subspaces which are estimated from spectral truncation. It reduces interference with historical representations without guaranteeing invariance of the complete nonlinear network. Future work could develop tighter theoretical analyses and task-aware subspace selection strategies.

Finally, CD relies on a fixed reference set and a frozen CLIP as teacher. Its effectiveness depend on the semantic coverage of the reference data, while the teacher cannot incorporate knowledge acquired from tasks. Promising directions include data-free distillation, multi-teacher learning, and extensions to multimodal large language models (MLLMs). For MLLMs, null-space projection could be applied to the visual projector and selected language-model layers, while token-level distillation could preserve pretrained cross-modal knowledge. Future studies should address the challenges posed by larger model scales, longer sequences, and continual multi-tasks learning.

\section{Acknowledgement}
Supported by the China Postdoctoral Science Foundation under Grant Numbers BX20240013 and 2024M760113, the Guangdong S\&T Program (2024B0101010003), the Shenzhen Science and Technology Innovation Commission under Grant KQTD 20240729102051063, and the National Natural Science Foundation of China under Grant Numbers U25B6003 and 62425101.

\bibliographystyle{IEEEtran}
\bibliography{main}

\end{document}